# Application of Multi-Core Parallel Programming to a Combination of Ant Colony Optimization and Genetic Algorithm

Rishita Kalyani
VIT University
Chennai, India
rishitapankaj.kalyani2011@vit.ac.in

*Abstract*—This Paper will deal with a combination of Ant Colony and Genetic Programming Algorithm to optimize Travelling Salesmen problem (NP-Hard). However, the complexity of the algorithm requires considerable computational time and resources. Parallel implementation can reduce the computational time. In this paper, emphasis in the parallelizing section is given to Multi-core architecture and Multi-Processor Systems which is developed and used almost everywhere today and hence, multi-core parallelization to the combination of algorithm is achieved by OpenMP library by Intel Corporation.

## 1. INTRODUCTION

Ant Algorithms were first proposed by Dorigo and colleagues [1, 4] to solve the various NP-Hard problems. This introduction was a breakthrough in the field of combinational optimization problems. Ant Colony Optimization (ACO) problem works on three parameters – α, β and δ. The importance of a trail and visibility are the factors determined by these parameters. These are the probability and the heuristic parameters of the algorithm. In this paper we try to find out the best result of a graph by passing the result acquired from ACO to the standard genetic algorithm and then optimize the computation time by using multi-core parallel programming. The first section will deal with standard ACO and standard genetic algorithm, then we will discuss how we have combined both to implement together to further optimize the result and then the last section discusses different ways to parallelize. Both these algorithms are construction algorithms, and hence, a solution is built on making use of some problem-specific heuristic information, ACO further extends this traditional construction heuristic with an ability to exploit the experience gathered during the optimization process. There is a lot of research going on in the ant algorithms; combined with various other natural algorithms like Genetic Algorithm (GA) and Particle Swarm Optimization (PSO) covering problems like vehicle routing, graph coloring, sequential ordering, routing in communication networks and many more.

### 2.1 THE STANDARD ANT COLONY OPTIMIZATION

Ant algorithms are inspired by the observations taken from real ant colonies. Ants are social insects [1]. They have a habitat maintained in colonies and their survival is entirely dependent on the survival of the entire colony. The foraging behavior of ants is studied to understand how they find the shortest path between food and their nest over a period of time. Ants form a pheromone trail while they forage. Other ants of the same colony can smell this pheromone deposited by previous ants and hence, determine the path by giving probability to the strength of pheromone deposited. This trail also allows ants to find their way back. Figure 1 explains how a typical ant colony will behave on encountering an obstacle.

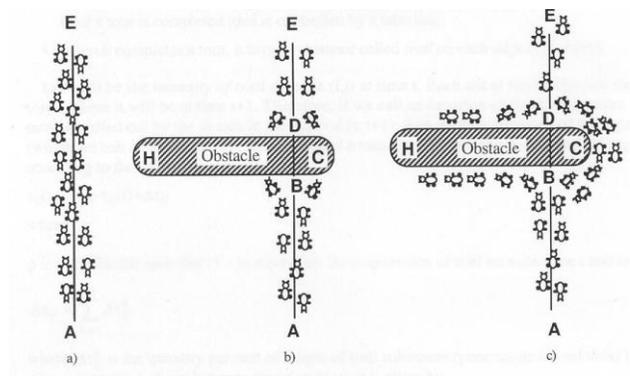

Figure 1: The ants travel from A (nest) to E (food) and then back to A. If they encounter any obstacle in between then all the ants will in beginning divide themselves and equal number of ants will travel from both paths – A-H-E and A-C-E and then back to nest. [6, 7, 8]

The ants will then deposit pheromones and depending upon the strength of this pheromone in each path, the shortest path is chosen in the end. Figure 2, explains this phenomenon in brief. Let us say that at each step, 30 ants go from A to B and 30 travel from E to D while retracing back to nest. Now, at the junction B, 15 ants travel in each path - B-H-D and B-C-D and all ants deposit say 1 unit of pheromone per unit time. Hence, the strength of pheromone per unit area is more in the path B-C-D after 1 unit of time is spent (t=1) and hence, more and more ants will choose this path, eventually leading to all ants choosing the shortest path.

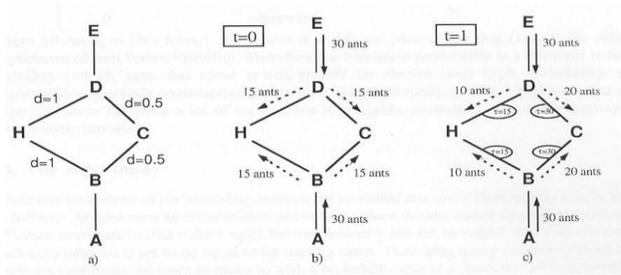

Figure 2: The strength of pheromone increases per unit area in B-C-D path after t=1.

The Travelling Salesmen Problem (TSP) is solved by ACO in the following manner. Let there be m ants placed on randomly chosen cities, then at each city state transition rule is applied in iterations. An ant at city $i$ choose to visit city $j$ which has probabilistic parameter – pheromone strength as $\tau_{ij}(t)$ on the arc and a locally available heuristic function which is information of arc length. Each ant has a restricted memory called *tabu* list in which current partial tours are stored. After every ant has constructed its tour, pheromones are updated. Pheromone trail strength is first lowered by a factor – evaporation factor and then the arcs visited by ants will have deposition of pheromone. $\tau_{ij}(t)$ is the learned suitability of choosing city $j$ from $i$. Let $\eta_{ij}$ be the visibility of $j$ and $i$ defined as

$$\eta_{ij} = \frac{1}{\text{distance}(i,j)} \quad (1)$$

Hence, bigger the product of $\tau$ and $\eta$, the better are the chances of $j$ being selected as the next city. The trail and visibility are represented by the parameters $\alpha$ and $\beta$. Therefore, the final formula to find the probability of choosing city j is as follows:

$$p_{ij}(t) = \begin{cases} [\tau_{ij}(t)]^{\alpha} \cdot [\eta_{ij}]^{\beta} / \sum_{l}[\tau_{il}(t)]^{\alpha} \cdot [\eta_{il}]^{\beta}, & \text{if } j \in \{N - tabu(k)\} \\ 0, & \text{otherwise} \end{cases}$$

where,
l belongs to {N – *tabu*(k)}        (2)
k represents k-th ant of m ants

Apart from the above given parameters, there is one evaporation parameter – $\delta$ which is generally a constant less than one to take care that there is no unlimited accumulation over the trail by decreasing the pheromone intensity over time.

In general, all ACO algorithms follow a particular scheme; the following pseudo code outlines general ACO.

**procedure** ACO_algorithm_for_TSP
   set parameters,
   initialize pheromone trails,
   Represent the problem by graph

**while** (condition != termination condition)
  for each ant do
    randomly select a starting node
  Repeat
    move to the next node according to node transition rules
  Until a tour is complete
  for each edge do
    Update the pheromone intensity using pheromone
    updating rules.
**end**
Output the global best tour.

## 2.2 THE STANDARD GENETIC ALGORITHM

Genetic algorithm is inspired from natural process of evolution which depends upon the survival of the fittest. This algorithm has five main stages – Generation of initial population, Evaluation of fitness of every individual, Selection of better genes among all, Cross-over between these "good" individuals and the last step is Mutation step to bring in diversity. If the result achieved is optimal then termination takes place or else, fitness evaluation, selection, cross-over and mutation steps are repeated iteratively.

In Travelling Salesmen Problem, initial population is created by giving an identifier to each city. One city is selected as starting city and rest all cities are non-visited cities. This is the initialization process. The next step is calculating fitness which is done by first calculating path length of each tour and then dividing an individual's path length by total path length. Then the selection operator is applied to every participating individual followed by the assignment of probability to each individual for being chosen in the crossover operation. In the mutation process, two genes are randomly replaced with each other and when the next population is produced, it is decided whether convergence will be achieved or not. The evaluation, selection, crossover and mutation will re-iterate until termination condition will be achieved.

The following pseudo code is followed while implementing the standard genetic algorithm:-

  Initialization   //Generate random population
  **while** (condition!=termination condition)
    Calculate_fitness
    sort            // according to chromosome fitness
    do_crossover
    do_mutation
    remove_duplicates
  **end**

The above mentioned algorithm is used in optimizing the process of ACO.

## 3. THE ACO-GA ALGORITHM

This section, we propose an algorithm which uses Genetic process to enhance the result and performance of Ant Colony Optimization. This consists of two parts – initialization of values and the main part.

The second part of algorithm requires a best tour check; this part uses genetic process to optimize the ACO.

Psuedo code for ACO-GA algorithm is as follows:

**Initialization**
   Set_initial_parameters: variables, states, input trajectory
   Set_initial_pheromone_trail_value
   Set_starting_city_of_m_ants:
     initial_memory_of_each_ant=NULL

**while** (condition!=termination condition)
   for each ant do
     Repeat
      move to the next node according to node transition rules
     Until a tour is complete
   Apply Local Search
// Best Tour check – main part
   for each edge do
     Update()
     Evaporate_fixed_proportion_of_pheromone
     Perform_ant_cycle_update_of_pheromone
   for each ant performing ant cycle
     get best tour
   Create_new_population //based on pheromone trails
     Calculate_fitness
     sort /*according to chromosome fitness, assign probability*/
     do_crossover
     do_mutation
     remove_duplicates

**end**

The Genetic algorithm is applied at the last stage while creating new population based upon pheromone trail. This operation is applied to each individual selected from the population with a probability assigned based on fitness. It can be seen that the above code is a mere combination of ACO and GA. The first part is a typical ACO algorithm and the result achieved from it is passed to GA and then before re-iterating in ACO, GA produces a better population for re-iteration.

The need for GA is that Ant colony may produce redundant states and it is more efficient to minimize the graph to enhance the output behavior. Thus, each ant of the colony undergoes genetic operation and brings out a better graph for subsequent cycles.

### 3. PARALLELIZATION OF ALGORITHM

The ACO-GA algorithm discussed in this paper can be parallelized by multi-core parallelization. OpenMP programming is used to parallelize this algorithm. The parallelization can be achieved in various methodologies, i.e., by Threading Building Blocks parallelization or CUDA programming parallelization, however, we concentrate on OpenMP multi-core parallel programming here.

OpenMP API is a portable parallel programming model for shared memory multi-processor or multi-core architecture. It consists of compiler directives which allow the user to explicitly declare and define parallelism. OpenMP is basically used to convert a given sequential program into parallel; it exploits the parallelization of loop-based data and the incrimination in the speed of computation is solely dependent on the number of cores or processors.

In our algorithm, we consider the process of each ant searching the new state as independent process, since this involves no interference; we can easily parallelize this step. However, before we move on to the next step of applying algorithm of genetic operation, we need the result from each ant as we evaluate and remove all the redundant states from the graph. Therefore, we wait until all the threads from the block of ACO are computed and then the result is passed to the last section of program, which is application of Genetic algorithm.

**while** (condition!=termination condition)
*#pragma omp parallel for num_threads(num_size)*
   for each ant do
     Repeat
      move to the next node according to node transition rules
     Until a tour is complete
   Apply Local Search
// Best Tour check – main part
   for each edge do
     Update()
     Evaporate_fixed_proportion_of_pheromone
     Perform_ant_cycle_update_of_pheromone
   for each ant performing ant cycle
     get best tour

/*Now, before continuing, we force all the threads to complete their operations*/

*#pragma omp barrier*
   Create_new_population //based on pheromone trails
     Calculate_fitness
     sort /*according to chromosome fitness, assign probability*/
     do_crossover
     do_mutation
     remove_duplicates
**end**

The above program has one detail that needs to be specified here, which being that the definition of all the variables outside the parallel directive and inside the directive are different. All variables outside directive are shared variables for all threads. Hence, variables exclusively for k-th ant should be declared inside the thread or there could be a possibility where other thread might hinder with the computation of that variable and change the value of it. The step of declaring the number of threads (omp_set_num_threads) is optional. The user can set the number of threads as the number of cores in the system or let the compiler itself general n – a random number of threads divided among the cores of user's system.

## 4. CONCLUSION

The given parallelized ACO-GA algorithm enhances the behavior of the system and reduces the computational time to a certain extent. The computational time is reduced by two factors:
(i) Minimizing redundant states by passing the result achieved in ACO section of algorithm and hence, augments the behavior of the system.
(ii) Parallelizing the ACO section of the algorithm by multi-core parallel programming and hence, reducing the computation time considerably.

When an ant completes a solution or is in construction phase, the ant will evaluate the solution and will modify the trail values on the components which are used in the solution. Depending on these values, ants deposit a certain quantity of pheromone on the components (vertices or edge) and thereby, increasing or decreasing the probability of including it in the graph. The ants following will use this pheromone information to guide in achieving optimal and better solutions. The Genetic algorithm allows the evolution of system which can perform alternative computations which are conditioned upon the outcome of the intermediate results of ACO and hence, define and change the components to be used for the next subsequent iterations.

Parallelizing the entire algorithm by parallel programming with the help of OpenMP library, we exploit the rapid development of multi-core architecture.

## 5. FUTURE WORK

Ant colony optimization and Genetic algorithms are fairly new algorithms. They hold a possibility of quite a great deal of research to further optimize it. The factors determining the parameters of ACO which determines the importance of trail and visibility needs some research. Optimum values of these parameters will bring in optimum results. Also ACO and GA can be further combined with Particle Swarm Optimization algorithm to bring in a more promising result with optimum usage of resources and time.